\documentclass[a4paper,conference]{IEEEtran}
\usepackage{graphicx}
\usepackage{booktabs}
\usepackage{xcolor} 
\usepackage{hyperref}
\ifCLASSINFOpdf
\else
\fi

\newcommand{\ifRevesion}{\iftrue}
\newcommand{\ifAppendix}{\iffalse}
\newcommand{\sout}[1]{\ifRevesion \MS \sout{#1} \fin \fi}

\title{TocBERT: Medical Document Structure Extraction Using Bidirectional Transformers}

\author{\IEEEauthorblockN{Sarra Baghdadi\IEEEauthorrefmark{1}\IEEEauthorrefmark{2},
		Majd Saleh\IEEEauthorrefmark{1}\IEEEauthorrefmark{2} and
		Stéphane Paquelet\IEEEauthorrefmark{1}}
	\IEEEauthorblockA{\IEEEauthorrefmark{1}Institute of Research and Technology b-com, Rennes, France\\ 
		Sarra.BAGHDADI@b-com.com, Majd.SALEH@b-com.com, Stephane.PAQUELET@b-com.com}
	\IEEEauthorblockA{\IEEEauthorrefmark{2}Equal contribution}}

\date{\today}

\begin{document}
	
\maketitle

\begin{abstract}
Text segmentation holds paramount importance in the field of Natural Language Processing (NLP).
It plays an important role in several NLP downstream tasks like information retrieval and document summarization.
In this work, we propose a new solution, namely TocBERT, for segmenting texts using bidirectional transformers.
TocBERT represents a supervised solution trained on the detection of titles and sub-titles from their semantic representations.
This task was formulated as a named entity recognition (NER) problem.
The solution has been applied on a medical text segmentation use-case where the Bio-ClinicalBERT model is fine-tuned to segment discharge summaries of the MIMIC-III dataset.
The performance of TocBERT has been evaluated on a human-labeled ground truth corpus of $250$ notes. 
It achieved an F1-score of $84.6\%$ when evaluated on a linear text segmentation problem and $72.8\%$ on a hierarchical text segmentation problem. It outperformed a carefully designed rule-based solution, particularly in distinguishing titles from subtitles.
\end{abstract}

\renewcommand{\IEEEkeywordsname}{Keywords}

\IEEEpeerreviewmaketitle

\section{Introduction}
Text segmentation is the task of partitioning a document into topically coherent segments \cite{TxtSeg_word_embed, TxtSeg_stat, TxtSeg_2lvl_Transformer_sup_CATS, TxtSeg_2lvl_BiLSTM_sup}. It represents an important pre-processing step in several Natural Language Processing (NLP) downstream tasks inducing information retrieval and summarization\cite{TxtSeg_word_embed, TxtSeg_stat, TxtSeg_2lvl_Transformer_sup_CATS, TxtSeg_2lvl_BiLSTM_sup,Doc_sum_TxtSeg,LLM_IR_Survey}. Two types of text segmentation can be distinguished: 1) linear segmentation where a document is divided into sequential contiguous segments, and 2) hierarchical segmentation where higher level segments are further segmented into smaller sub-segments \cite{TxtSeg_2lvl_Transformer_sup_CATS}. 

Existing solutions to text segmentation are applied either to unformatted, i.e. plain-text, documents like \cite{TxtSeg_2lvl_Transformer_sup_CATS,TxtSeg_2lvl_BiLSTM_sup} or to formatted ones like \cite{FinTOC-2,FinTOC_2021,FinTOC_2022}. In the first family, recent text segmentation methods are based on the semantic representation of words and/or of sentences. For instance, word embeddings and sentence embeddings are used to detect topic changes in a sequence of sentences in \cite{TxtSeg_2lvl_Transformer_sup_CATS,TxtSeg_2lvl_BiLSTM_sup}.
In the second family, document layout, orthographic features, geometric features and stylistic properties, e.g. font properties and line spaces, are exploited to extract the document structure. Some of the aforementioned features can be extracted using the Document Image Analysis (DIA) \cite{DIA,FinTOC_2022} while others can be extracted using regex matches of predefined patterns \cite{FinTOC_2021}. In both families, titles and subtitles play an important role in text segmentation. For example, in semantic-aspects-based  methods, supervised solutions like \cite{TxtSeg_2lvl_Transformer_sup_CATS} and \cite{TxtSeg_2lvl_BiLSTM_sup} are trained on a large corpus named WIKI-727K \cite{TxtSeg_2lvl_BiLSTM_sup} which is automatically labeled such that topic-change cutoff points are completely dependent on titles. Similarly, visual-aspects-based methods like \cite{FinTOC_2022} are mainly based on title detection.
\begin{figure}[t]
	\centering
	\includegraphics[width=\columnwidth]{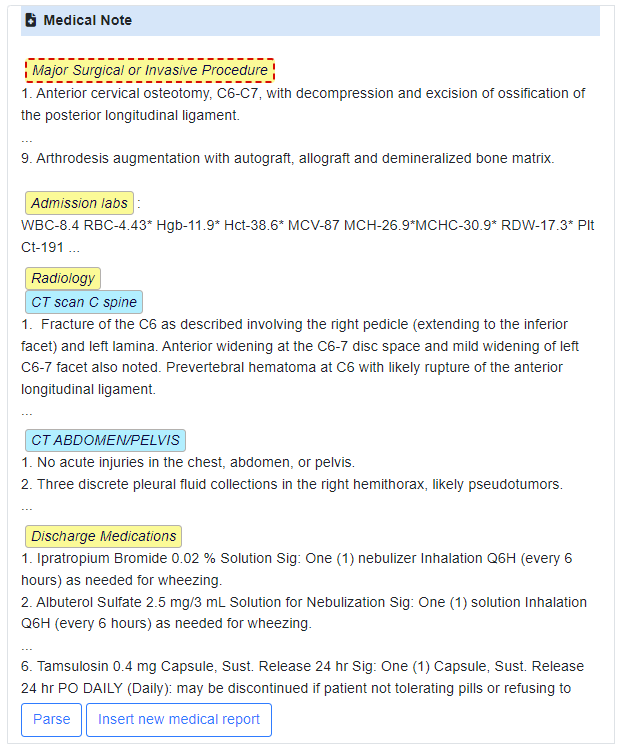}
	\caption{An extract from a discharge summary from the MIMIC-III database annotated with titles (yellow) and subtitles (blue)}
	\label{fig:example of titles of MIMIC-III}
\end{figure}

In this article, the focus is set on the \underline{hierarchical} segmentation of \underline{unformatted} medical documents by detecting titles and subtitles using their semantic vector representations. Particularly, the considered use-case is the segmentation of the discharge summaries of the MIMIC-III database \cite{MIMIC-III} which represents an unformatted free-text corpus. Figure \ref{fig:example of titles of MIMIC-III} shows an extract of a discharge summary where we highlighted titles in yellow and subtitles in blue. The motivation underlying the aforementioned segmentation task is two-fold: 
\begin{itemize}
	\item To clean the MIMIC-III discharge summaries corpus such that it can be efficiently used to domain-adapt pretrained language models. For example, noisy sections like "Discharge medications" or "Admission labs" can be detected and removed since their content doesn't represent a fluent language and can harm the domain adaption task.
	\item To enable the extraction of specialized sub-corpora and facilitate the information retrieval tasks. For instance, sub-corpora about \textit{radiology reports} or \textit{cardiac diseases reports} can be extracted and used to train relevant information extraction systems.
\end{itemize}
While the aforementioned advantages have been discussed on the considered use-case, the proposed method is general and can be extended to solve many similar tasks as will be shown in Section \ref{Sec:proposed_method}.

The rest of the paper is organized as follows: we start by exploring related works in Section \ref{Sec:related_work}. Training and test corpora are described in Section \ref{Sec:data}. The proposed method is presented in Section \ref{Sec:proposed_method}. Experimental results are discussed in Section \ref{Sec:results} while Section \ref{Sec:conclusion} concludes the paper and discuses future work.

\section{Related work} \label{Sec:related_work}

\begin{figure}[t]
	\centering
	\includegraphics[width=\columnwidth]{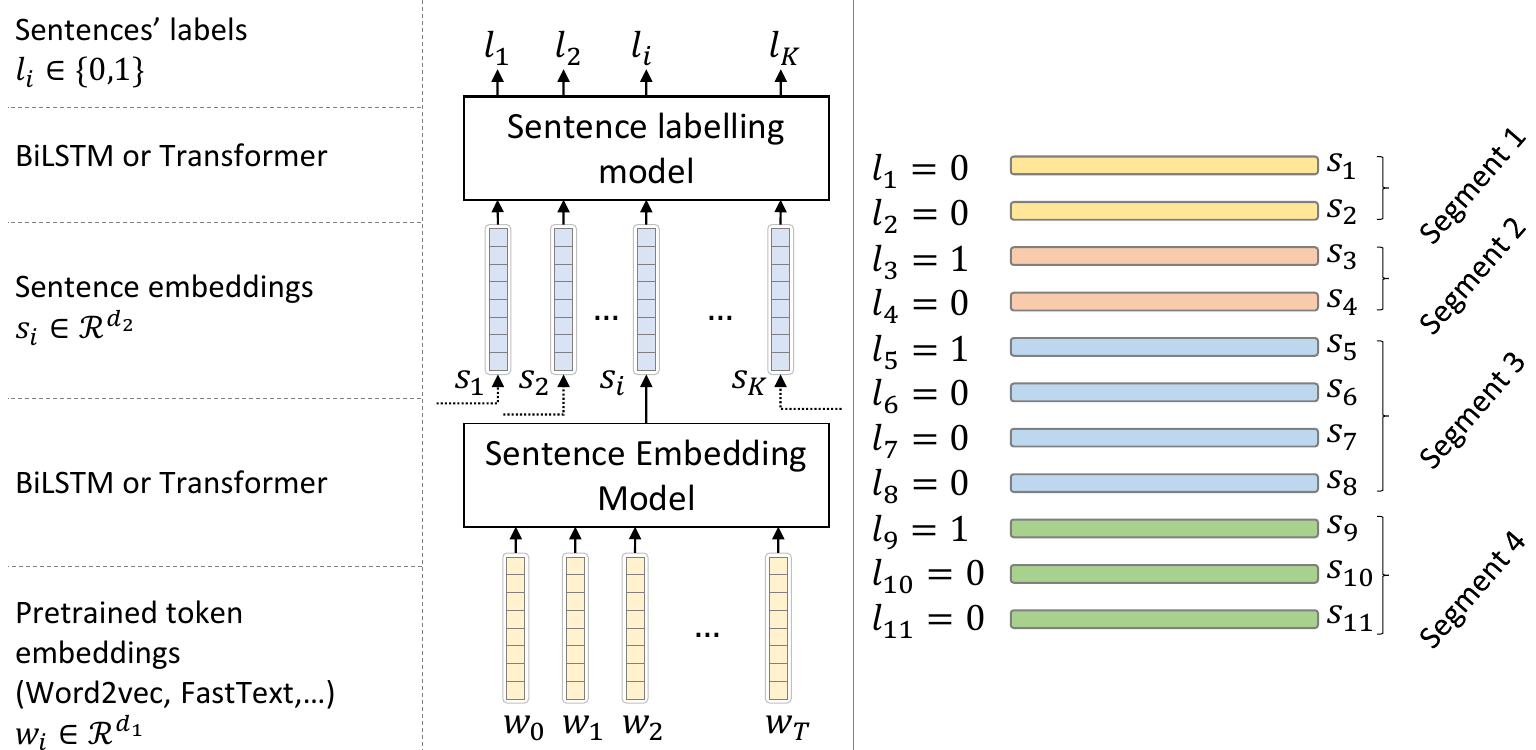}
	\caption{Two-level neural networks for text segmentation: the first-level network builds sentence embeddings from token embeddings while the second-level network labels the sequence of sentence representations as: topic-change $(1)$ or no-topic-change $(0)$}
	\label{fig:Txt_2lvl}
\end{figure}

One remarkable family of solutions to topical text segmentation was proposed in \cite{TxtSeg_2lvl_BiLSTM_sup,TxtSeg_2lvl_Transformer_sup_CATS}. As depicted in Figure \ref{fig:Txt_2lvl}, these methods are based on two-level neural networks that divide the segmentation problem into two sub-problems: 1) starting from token embeddings, build sentence embeddings, and 2) based on the sentence representations, annotate the sentences' sequence with binary labels i.e. $0$ for no-topic-change and $1$ for topic-change. Specifically, these methods start by dividing the input document into sentences (sentence tokenization) using the NLTK PUNKT tokenizer \cite{PUNKT_tokenizer}. Then, they use pretrained token embeddings to represent input tokens $\textbf{w}_i$. Concretely, word2vec embeddings \cite{word2vec1,word2vec2} are used in \cite{TxtSeg_2lvl_BiLSTM_sup} while FastText embeddings \cite{fasttext} are used in \cite{TxtSeg_2lvl_Transformer_sup_CATS}. Token embeddings of each sentence are fed to a sentence embedding model in order to create sentence representations i.e. each sentence is represented by a dense vector $\textbf{s}_j$. This sentence embedding model is a 2-layer bidirectional LSTM \cite{LSTM} in the method proposed in \cite{TxtSeg_2lvl_BiLSTM_sup} while it represents a transformer encoder \cite{Transformers} in the method proposed in \cite{TxtSeg_2lvl_Transformer_sup_CATS}. The sequence of sentence embeddings is then fed to a sentence labeling model which assigns a binary label $l_i \in\{0,1\}$ to each sentence where $l_i=1$ encoding the existence of a topic change. This sentence labeling model is a 2-layer bidirectional LSTM in \cite{TxtSeg_2lvl_BiLSTM_sup} while it is a transformer encoder in \cite{TxtSeg_2lvl_Transformer_sup_CATS}. Both of the aforementioned solutions are trained in a supervised manner where the labeled corpus WIKI-727K \cite{TxtSeg_2lvl_BiLSTM_sup} is used for training. Note that the transformer-based solution adds another feed-forward head besides the sentence labeling model. This head is responsible for explicit coherence modeling \cite{TxtSeg_2lvl_Transformer_sup_CATS}.
A similar work was proposed in \cite{TxtSeg_Cross_Seg_Attention} where authors tested a solution based on a transformer to encode sentences (first level) and a bidirectional LSTM to detect topic changes (second level).
For deeper details about how transformers and LSTMs work, interested readers are referred to this tutorial \cite{AnatomyNLM}.

While the aforementioned solutions are supervised, many solutions based on unsupervised-learning have been proposed. For instance, the method proposed in \cite{BERT_Topic_seg_nsup} uses RoBERTa embeddings \cite{RoBERTa} to calculate semantic similarity scores between sentences. These scores are then used to perform segmentation. Particularly, the variation in similarity scores over time is leveraged to recognize changes in topics. The problem of representing sentences from token embeddings was addressed in two ways: 1) to use sentence-BERT \cite{SBERT}, a version of BERT \cite{BERT} fine-tuned on sentence representation tasks, and 2) to aggregate token embeddings from the 2nd to the last layer of RoBERTa using max pooling.

Before the deep-learning era, several interesting statistical solutions were proposed. Interested readers are referred for example to 1) TextTiling \cite{TextTiling} which captures semantic similarity between sentences using word frequency vectors, 2) TopicTiling \cite{TopicTiling} an algorithm based on TextTiling \cite{TextTiling} and Latent Dirichlet Allocation (LDA) \cite{LDA}, 3) LDA-based topic modeling \cite{TxtSeg_topic_models_LDA}, and 4) the statistical models for text segmentation proposed in \cite{TxtSeg_stat}.

The solutions discussed so far represent the \textit{semantic-aspects-based} text segmentation family. Let us explore now some \textit{visual-aspects-based} text segmentation methods i.e. segmentation applied on formatted documents. In fact, in the financial domain, several interesting solutions to the document-structure-extraction problem have been proposed thanks to the FinTOC shared tasks \cite{FinTOC_2021,FinTOC-2,FinTOC_2022}. These latter aim at extracting the structure of complex layout formatted documents stored in pdf format. The objective is to detect titles and to hierarchically organize them into a Table of Contents (ToC). For instance, authors of \cite{FinTOC_2022} proposed a title detection algorithm based on Document Image Analysis (DIA). Particularly, they started by Faster R-CNN \cite{Faster_R_CNN} model which is pertained on the PubLayNet dataset \cite{PubLayNet} and they fine tuned it on the FinTOC-2022 training set. In order to recognize the hierarchical levels of the detected titles, they extracted orthographic and layout features from the detected titles and they proposed a supervised random forest model to predict the title level.

\section{Data analysis and annotation} \label{Sec:data}
\begin{figure}[t]
	\centering
	\includegraphics[width=\columnwidth]{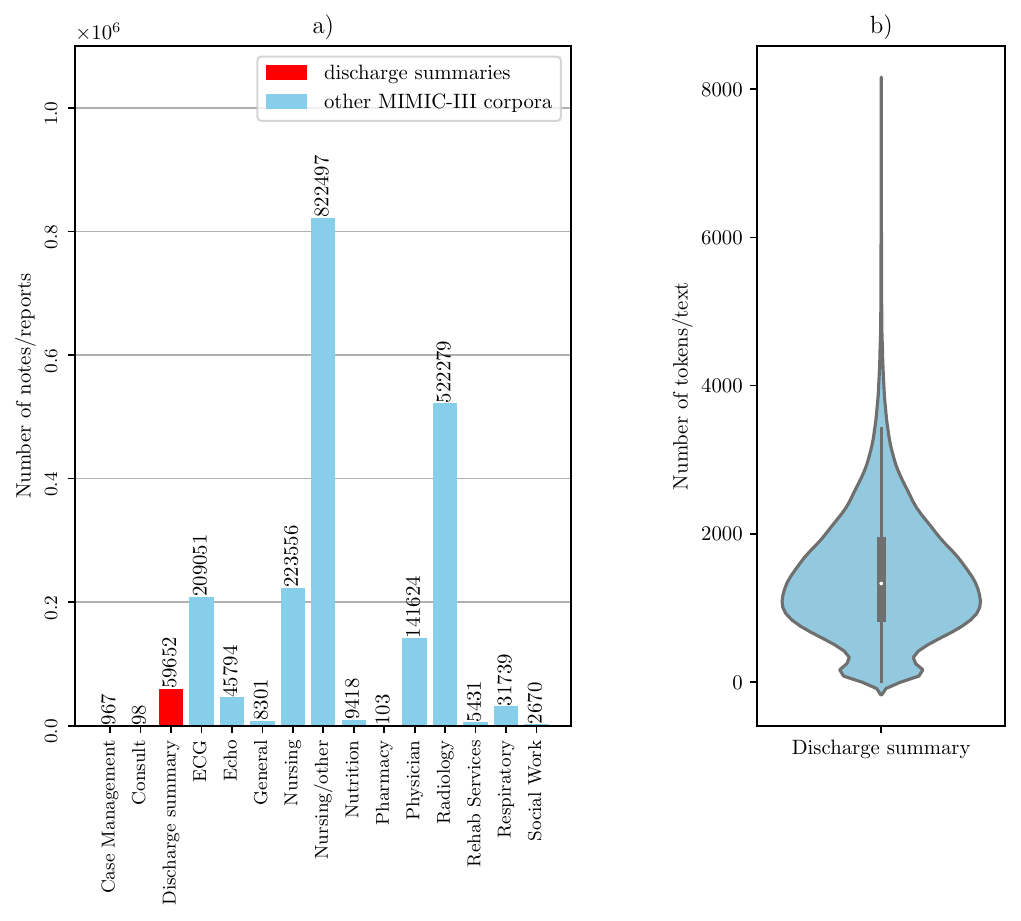}
	\caption{a) distribution of sub-corpora sizes of the MIMIC-III database. 2) distribution of reports' lengths of the discharge summaries corpus.
	}
	\label{fig:MIMIC3_stat}
\end{figure}

The experimental work has been performed on the discharge summaries corpus, a sub-corpus from the MIMIC-III \cite{MIMIC-III} database. MIMIC-III contains more than two million reports including discharge summaries, EEG reports, radiology reports, and many others. Figure \ref{fig:MIMIC3_stat}-a shows the distribution of MIMIC-III reports' families. The discharge summaries corpus contains around $60~000$ reports. Figure \ref{fig:MIMIC3_stat}-b shows the distribution of discharge summaries' lengths given in number of tokens. The average discharge summary length is $1435$ tokens (roughly $3$ pages).

We divided the discharge summaries corpus to a test corpus of $250$ randomly selected reports and a training corpus containing the rest of reports. The test corpus has been manually annotated using the web application ACUITEE \cite{ACUITEE_software} as illustrated in Figure \ref{fig:example of titles of MIMIC-III}. Note that two hierarchical level of titles are considered: titles (highlighted in yellow) and sub-titles (highlighted in blue). 

In order to annotate the training corpus, we conducted a deep analysis of the discharge summaries structure and we proposed a rule-based title detection system, namely TocRegex. This latter has been implemented using regular expressions. For example, the first considered pattern was any character sequence that:
\begin{itemize}
	\item \textit{begins with} a newline "\textbackslash n" followed by a valid title content and \textit{terminates with} a colon followed by newline ":\textbackslash n"; or
	\item \textit{begins with} a double newline "\textbackslash n\textbackslash n" followed by a valid title content and \textit{terminates with} a colon ":"; or
	\item \textit{begins with} a beginning-of-document character followed by a valid title content and \textit{terminates with} a colon ":"
\end{itemize}
The definition of "valid title content" as well as the full list of the $12$ considered regular expressions are discussed in the Appendix \ref{App:TocRegex}.\\
Figure \ref{fig:regex_title_freq} shows the frequencies of the top 35 matches detected using the above-described pattern. For instance, the title "history of present illness" has been detected $10e5$ times in the corpus. We note that several false positives have also matched the considered pattern like e.g. "tablet(s)*refills" which appears frequently in medication lists. In order to filter-out such kind of false positives, a manual curation has been applied to the detected titles. As a result, the total remaining numbers of unique titles is $56~357$. They have appeared $2~930~725$ times in the corpus i.e. the average number of titles is around $49$ per discharge summary.

\begin{figure*}[t]
	\centering
	\includegraphics[width=\textwidth]{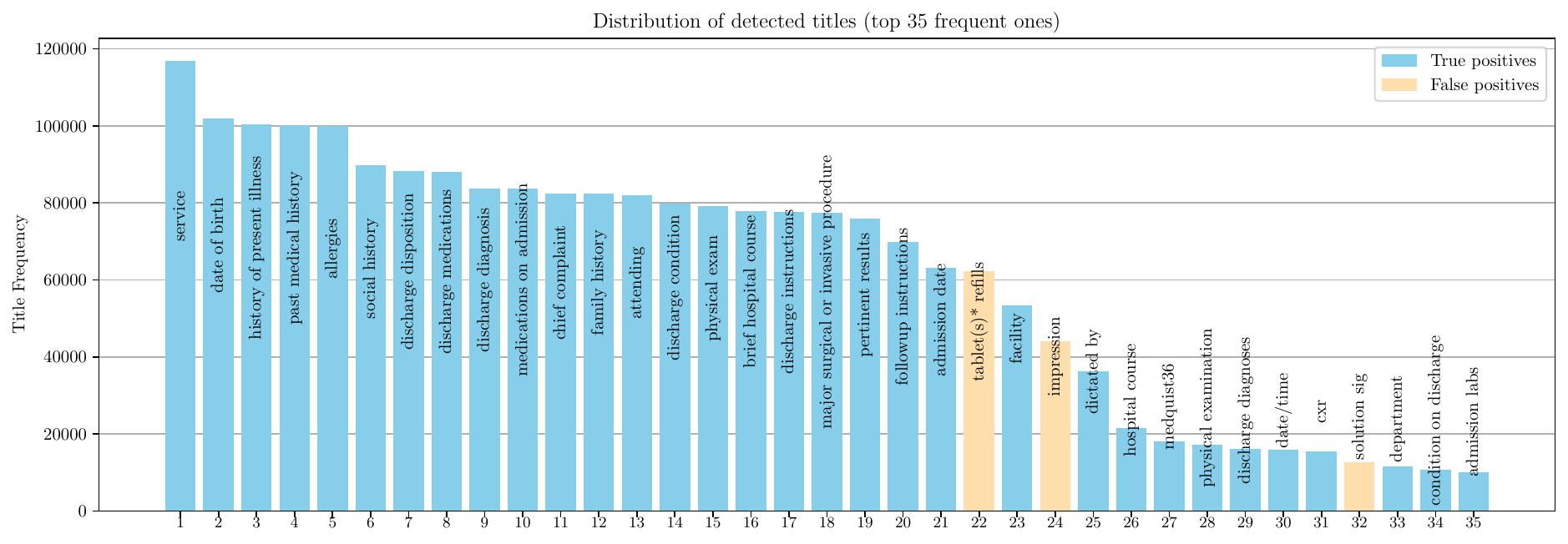}
	\caption{Top $35$ "candidate" titles extracted using the first pattern of the TocRegex solution including true positives (blue) and false positives (light brown)
	}
	\label{fig:regex_title_freq}
\end{figure*}

\section{The proposed solution: TocBERT} \label{Sec:proposed_method}
\begin{figure*}[h]
	\centering
	\includegraphics[width=\textwidth]{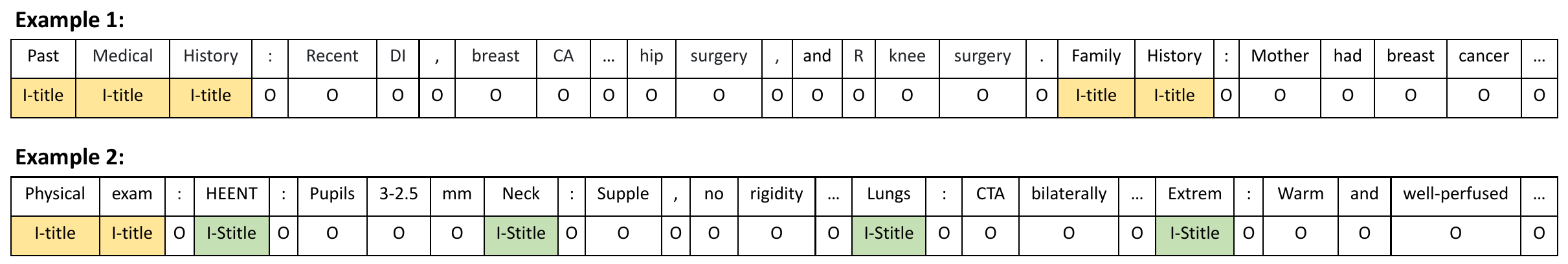} 
	\caption{Labeling token sequences in TocBERT: Example 1) two titles "Past Medical History" and "Family History" are labeled with "I-title" while other tokens are labeled with "O"; Example 2) one title "Physical exam" is labeled with "I-title" and four subtitles "HEENT", "Neck", "Lungs" and "Extremities" are labeled with "I-Stitle", while other (outside) tokens are labeled with "O".}
	\label{fig:example of labeled data}
\end{figure*}

In this section, we explore the proposed solution TocBERT (Table of Content BERT). The hierarchical title detection task is formulated as a sequence labeling problem i.e. token classification problem. Particularly, it is considered as a named entity recognition (NER) problem with three entity types: "I-title" for titles, "I-Stitle" for subtitles and "O" for other tokens. Note that we adopt the "Inside–outside–beginning" (IOB) standard labeling format. Figure \ref{fig:example of labeled data} shows two examples of labeled sequences where tokens are tagged with \textit{title}, \textit{subtitle} or \textit{outside} labels. For instance, the second sequence has one title "Physical exam" and four subtitles "HEENT", "Neck", "Lungs" and "Extremities".

TocBERT is based on fine-tuning the pretrained model Bio-ClinicalBERT \cite{Bio-ClinicalBERT} on the above-mentioned NER task.
Note that bidirectional transformers, like the BERT family, are more convenient for tackling sequence labeling tasks, like NER, compared to generative i.e. auto-regressive transformers like GPT \cite{GPT1}.
Bio-ClinicalBERT represents a variant of BERT adapted to biomedical and clinical domains. We fine-tuned this pretrained model using the discharge summaries training corpus that has been semi-automatically labeled using TocRegex followed by a manual curation as described in the previous section \ref{Sec:data}. \\
Bio-ClinicalBERT, like BERT, is pretrained using a fixed size vocabulary: $28~996$ tokens. This vocabulary was created using the word-piece algorithm \cite{word_piece}. Some of the resultant tokens represent sub-words. For this reason, the first step of TocBERT is to project labels of the full-word tokens (extracted by simple pre-tokenization procedure) to sub-word tokens (extracted using Bio-ClinicalBERT tokenizer). 
The second step consists in preparing convenient training windows from the training corpus. This is important because the maximum window size of BERT is $512$ (sub-word) tokens while the average length of a discharge summary is $1435$ (full-word) tokens. To this end, discharge summaries have been segmented into windows of $384$ words where we used the approximate formula: $1$ token $= 0.75$ words. The total size of the resultant training set is $144~000$ labeled windows. 
Finally, the last step consists of training the TocBERT model. 

\section{Experimental results} \label{Sec:results}

\subsection{Experimental configurations}
TocBERT was trained on one NVIDIA A100 GPU equipped with 80 GB of RAM. It was trained for $20$ epochs with a batch size of $16$ training samples. The training time was around $17$ hours.
The inference hardware was an NVIDIA RTX A3000 laptop GPU equipped with 6 GB of dedicated RAM.

\subsection{Results}
Table \ref{tab:results_hierarchical} shows the experimental results of the proposed text segmentation solutions, TocBERT and TocRegex, in the hierarchical segmentation configurations, i.e. detecting both titles and sub-titles. While TocBERT is trained on a corpus labeled using TocRegex, the former considerably outperforms the latter in all the considered performance criteria, i.e. precision, recall and F1-score.

\begin{table}[h] 
	\centering 
	\caption{Experimental results: hierarchical text segmentation} 
	\label{tab:results_hierarchical} 
	\begin{tabular}{lcccccc} 
		\toprule 
		& Precision & Recall & F1-score \\ 
		\midrule 
		TocBERT  & \textbf{0.714} & \textbf{0.754} & \textbf{0.728}  \\ 
		\midrule 
		TocRegex  & 0.667 & 0.563 & 0.606  \\ 
		\bottomrule 
	\end{tabular}	
\end{table}

Table \ref{tab:results_linear} shows the results of the aforementioned solutions in the linear text segmentation configurations. TocBERT and TocRegex show comparable overall performance measured by the F1-score metric. TocBERT shows higher sensitivity (recall) while it is less specific. Note that TocBERT is completely based on the semantic representation of titles and their context. It doesn't depend on the existence of visual aspects like newlines and colons. This explains why it detects more titles (higher recall). On the other hand, TocRegex is more specific since it is entirely based on patterns that almost always exist only in titles.

Comparing the results of hierarchical and linear segmentation, the value of TocBERT is clearly its strong capacity to exploit the context and the semantic aspects to discriminate between titles and sub-titles. 

\begin{table}[h] 
	\centering 
	\caption{Experimental results: linear text segmentation} 
	\label{tab:results_linear} 
	\begin{tabular}{lcccccc} 
		\toprule 
		& Precision & Recall & F1-score \\ 
		\midrule 
		TocBERT  & 0.829 & \textbf{0.878} & \textbf{0.846}  \\ 
		\midrule 
		TocRegex  & \textbf{0.932} & 0.784 & 0.845  \\ 
		
		\bottomrule 
	\end{tabular}	
\end{table}

\begin{table*}[h] 
	\centering 
	\caption{Regular expressions used to detect titles} 
	\label{tab:regex_patterns} 
	\begin{tabular}{c} 
		\raisebox{-\totalheight}{\includegraphics[width=0.92\textwidth]{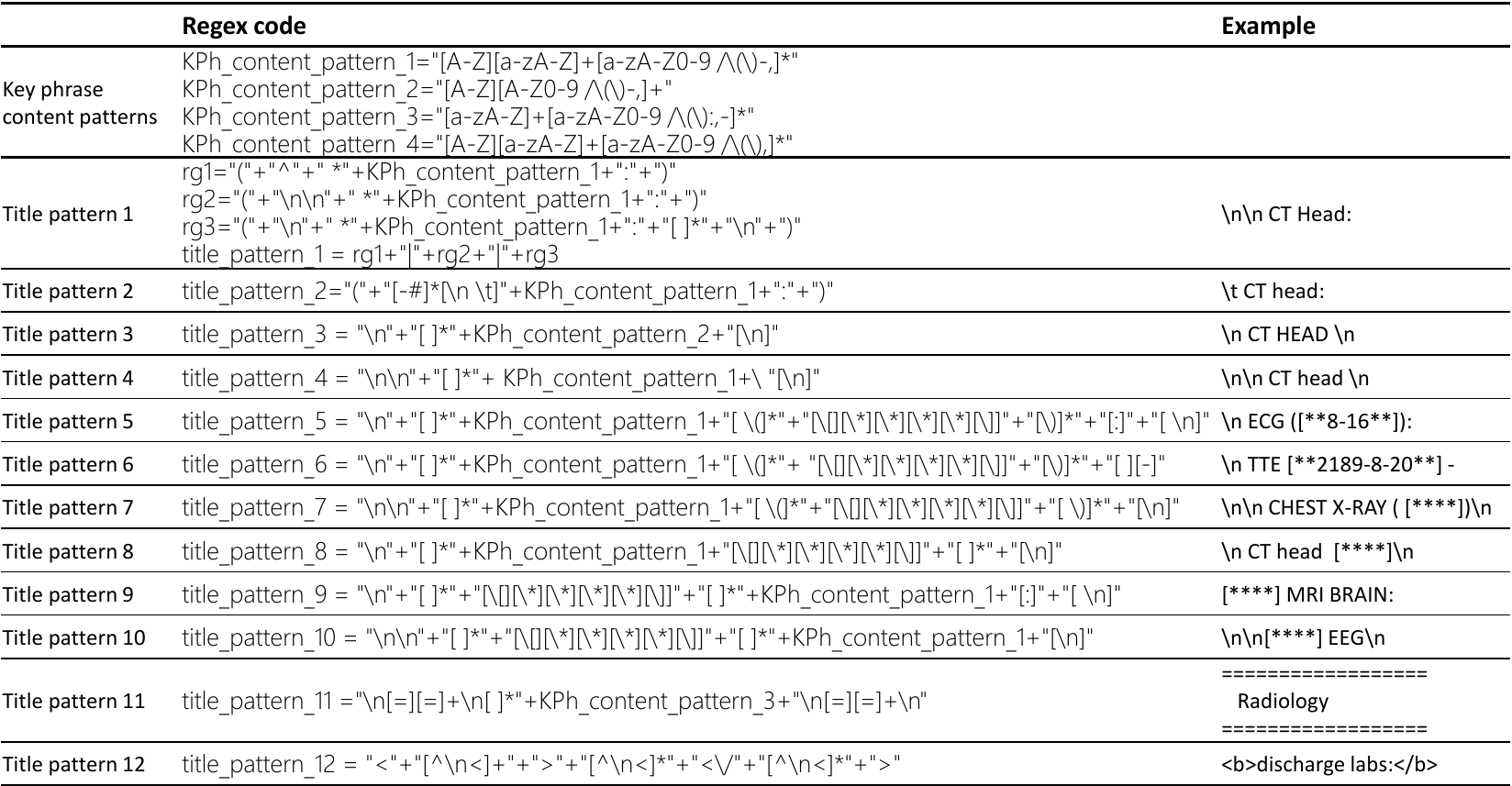}}
	\end{tabular}	
\end{table*}

In terms of inference execution time, TocRegex takes $87$ ms while TocBERT takes $195$ ms, in average, to segment a discharge summary.
While TocRegex is faster, both solutions satisfy near-real-time requirements.

\begin{figure}[h]
	\centering
	\includegraphics[width=\columnwidth]{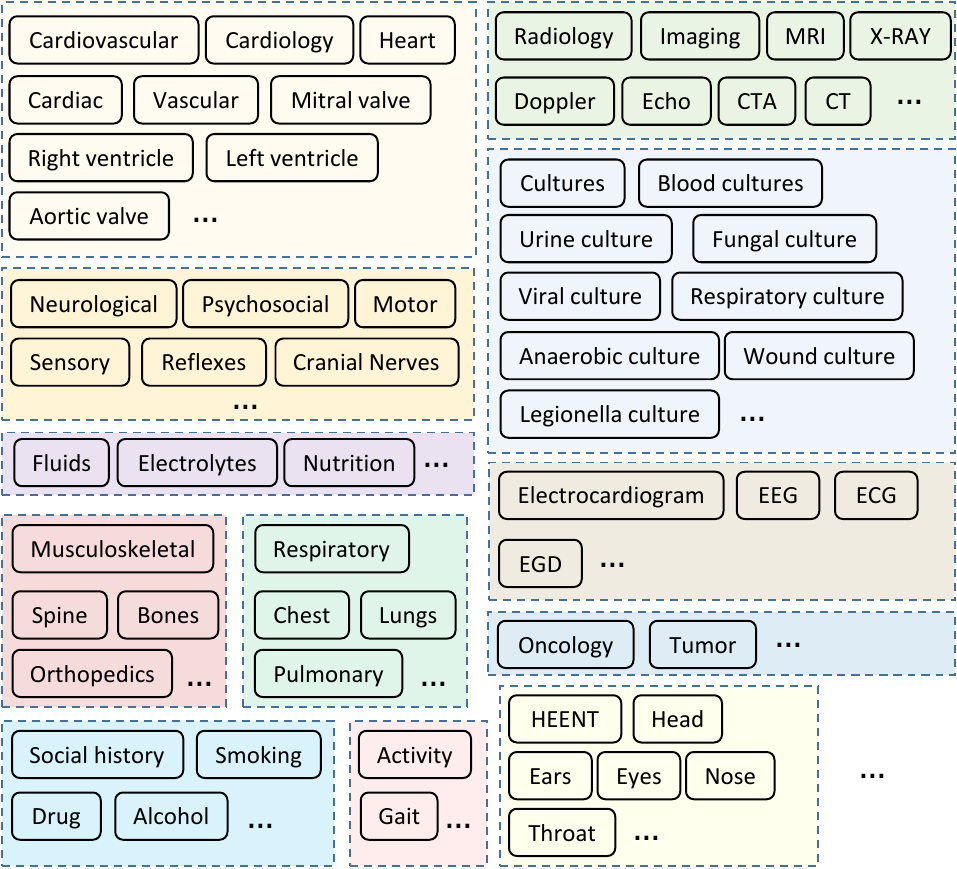} 
	\caption{Grouping of some key concepts towards constructing topics' ontology}
	\label{fig:ontology_example}
\end{figure}

Finally, we observed that the detected groups of titles and subtitles can be exploited to automatically construct an ontology of topics.
Figure \ref{fig:ontology_example} shows an example of initial manual grouping of key concepts that help constructing such ontology.
The semantic vector representations of the detected titles can play an important role in organizing the topics' ontology.
The interest of building this ontology is to facilitate tasks like text segmentation and information retrieval.
This idea will be investigated in future works.

\section{Conclusion} \label{Sec:conclusion}
In this paper, we proposed a new solution, TocBERT, for the hierarchical segmentation of medical reports. The segmentation task was formulated as a named entity recognition problem.  TocBERT was initialized by a pretrained model, Bio-ClinicalBERT, and fine tuned on a the MIMIC-III discharge summaries corpus. This latter was semi-automatically labeled with titles and sub-titles.
TocBERT showed very good results considerably outperforming a carefully-designed rule-based system. Particularly, it showed a good performance in discriminating between titles and subtitles by leveraging their semantic representations and employing their context.

The semantic representations of the extracted titles can be exploited to automatically construct an ontology of topics. Such an ontology can further facilitate tasks like text segmentation and information retrieval. The investigation of this idea is left for future work.

\appendices

\section{TocRegex} \label{App:TocRegex}
Table \ref{tab:regex_patterns} shows the regular expressions list used in the proposed rule-based solution, TocRegex. In the first line, the four supported title-content patterns are defined while the remaining $12$ lines list the full-title patterns.

\bibliographystyle{ieeetr} 

\bibliography{TocBERT}

\end{document}